\documentclass[sigconf, nonacm,pdfa]{acmart}

\newcommand\vldbdoi{10.14778/3632093.3632103}
\newcommand\vldbpages{386-399}
\newcommand\vldbvolume{17}
\newcommand\vldbissue{3}
\newcommand\vldbyear{2023}

\newcommand\vldbauthors{\authors}
\newcommand\vldbtitle{\shorttitle} 
\newcommand\vldbavailabilityurl{https://github.com/real2fish/CSL}
\newcommand\vldbpagestyle{empty}

\usepackage[a-2b]{pdfx}
\usepackage{multirow}
\usepackage{makecell}
\theoremstyle{plain}

\usepackage{amsmath}

\usepackage{amssymb}
\usepackage{mathtools}
\usepackage{cleveref}
\usepackage{url}

\usepackage{graphicx}

\usepackage{subcaption}
\usepackage{enumitem}

\usepackage{balance}

\newcommand{\xv}{\boldsymbol{x}}

\newcommand{\zv}{\boldsymbol{z}}

\newcommand{\eat}[1]{}

\newcommand{\myem}[1]{\textbf{\textit{#1}}}
\begin{document}
\title{TimeCSL: Unsupervised Contrastive Learning of General Shapelets for Explorable Time Series Analysis}

%%
%% The "author" command and its associated commands are used to define the authors and their affiliations.
\author{Zhiyu Liang}
\affiliation{%
  \institution{Harbin Institute of Technology}
  \city{Harbin}
  \country{China}  
}
\email{zyliang@hit.edu.cn}

\author{Chen Liang}
\affiliation{%
  \institution{Harbin Institute of Technology}
  \city{Harbin}
  \country{China}
}
\email{23B903050@stu.hit.edu.cn}

\author{Zheng Liang}
\affiliation{%
 \institution{Harbin Institute of Technology}
  \city{Harbin}
  \country{China}
}
\email{lz20@hit.edu.cn}

\author{Hongzhi Wang}
\affiliation{%
   \institution{Harbin Institute of Technology}
  \city{Harbin}
  \country{China}
  }
\email{wangzh@hit.edu.cn}\authornote{Corresponding author.}

\author{Bo Zheng}
\affiliation{%
   \institution{CnosDB Inc.}
  \city{Beijing}
  \country{China}
  }
  \email{harbour.zheng@cnosdb.com}


\begin{abstract}
Unsupervised (a.k.a. Self-supervised) representation learning (URL) has emerged as a new paradigm for time series analysis,  because it has the ability to learn generalizable time series representation beneficial for many downstream tasks without using labels that are usually difficult to obtain.  Considering that existing approaches have limitations in the design of the representation encoder and the learning objective, we have proposed Contrastive Shapelet Learning (CSL), the first URL method that learns the general-purpose shapelet-based representation through unsupervised contrastive learning, and shown its superior performance in several analysis tasks, such as time series classification, clustering, and anomaly detection. In this paper, we develop TimeCSL, an end-to-end system that makes full use of the general and interpretable shapelets learned by CSL to achieve explorable time series analysis in a unified pipeline. We introduce the system components and demonstrate how users interact with TimeCSL to solve different analysis tasks in the unified pipeline, and gain insight into their time series by exploring the learned shapelets and representation.   
\end{abstract}

\maketitle

\pagestyle{\vldbpagestyle}
\begingroup\small\noindent\raggedright\textbf{PVLDB Reference Format:}\\
\vldbauthors. \vldbtitle. PVLDB, \vldbvolume(\vldbissue): \vldbpages, \vldbyear.\\
\href{https://doi.org/\vldbdoi}{doi:\vldbdoi}
\endgroup
\begingroup
\renewcommand\thefootnote{}\footnote{\noindent
This work is licensed under the Creative Commons BY-NC-ND 4.0 International License. Visit \url{https://creativecommons.org/licenses/by-nc-nd/4.0/} to view a copy of this license. For any use beyond those covered by this license, obtain permission by emailing \href{mailto:info@vldb.org}{info@vldb.org}. Copyright is held by the owner/author(s). Publication rights licensed to the VLDB Endowment. \\
\raggedright Proceedings of the VLDB Endowment, Vol. \vldbvolume, No. \vldbissue\ %
ISSN 2150-8097. \\
\href{https://doi.org/\vldbdoi}{doi:\vldbdoi} \\
}\addtocounter{footnote}{-1}\endgroup

\ifdefempty{\vldbavailabilityurl}{}{
\vspace{.3cm}
\begingroup\small\noindent\raggedright\textbf{PVLDB Artifact Availability:}\\
The source code, data, and/or other artifacts have been made available at \url{\vldbavailabilityurl}.
\endgroup
}

\section{Introduction}\label{sec:intro}
A time series is one (a.k.a. univariate) or a group (multivariate) of variables observed over time. Time series analysis by discovering dependencies of the time-evolving variables can offer important insights into the underlying phenomena, which is useful for real-world applications in various scenarios, such as manufacturing~\cite{kim2019fault}, medicine~\cite{riley2021internet} and finance~\cite{bennett2022detection}. 

A major challenge in modeling time series is the lack of labels, because the underlying states required for labeling these time-dependent data are difficult to understand, even for domain specialists~\cite{TNC}. For this reason, recent studies focus on unsupervised (a.k.a. self-supervised) representation learning (URL) of time series~\cite{ts2vec,TS-TCC,TST,TNC,T-Loss,BTSF}, which aims to train a neural network (called encoder) without accessing the ground-truth labels to embed the data into feature vectors. The learned features (a.k.a. representation) can then be used for training models to solve a downstream analysis task, using \myem{little annotated data} compared to traditional supervised methods~\cite{TST}. Not only that, the features are more \myem{general-purpose} since they can benefit several tasks.

Unfortunately, existing URL approaches for time series have two limitations. First, these methods focus on the representation encoders based on the convolutional neural network (CNN)~\cite{causal_conv_wavenet,resnet} and the Transformer~\cite{attention}. However, these architectures are originally designed for domains such as computer vision and natural language processing, and have been shown to face many difficulties in modeling time series, due to the lack of capability to deal with the characteristics specific to time series ~\cite{OSCNN,FEDformer}. Second, some existing approaches are based on domain-specific assumptions. For example, ~\citet{T-Loss} and \citet{TNC} assume that the subseries distant in time are dissimilar, but it is easily violated in periodic time series~\cite{sanei2013eeg}. As a result, these methods cannot be well generalized to different scenarios.

\begin{figure}
    \centering
    \includegraphics[width=\linewidth]{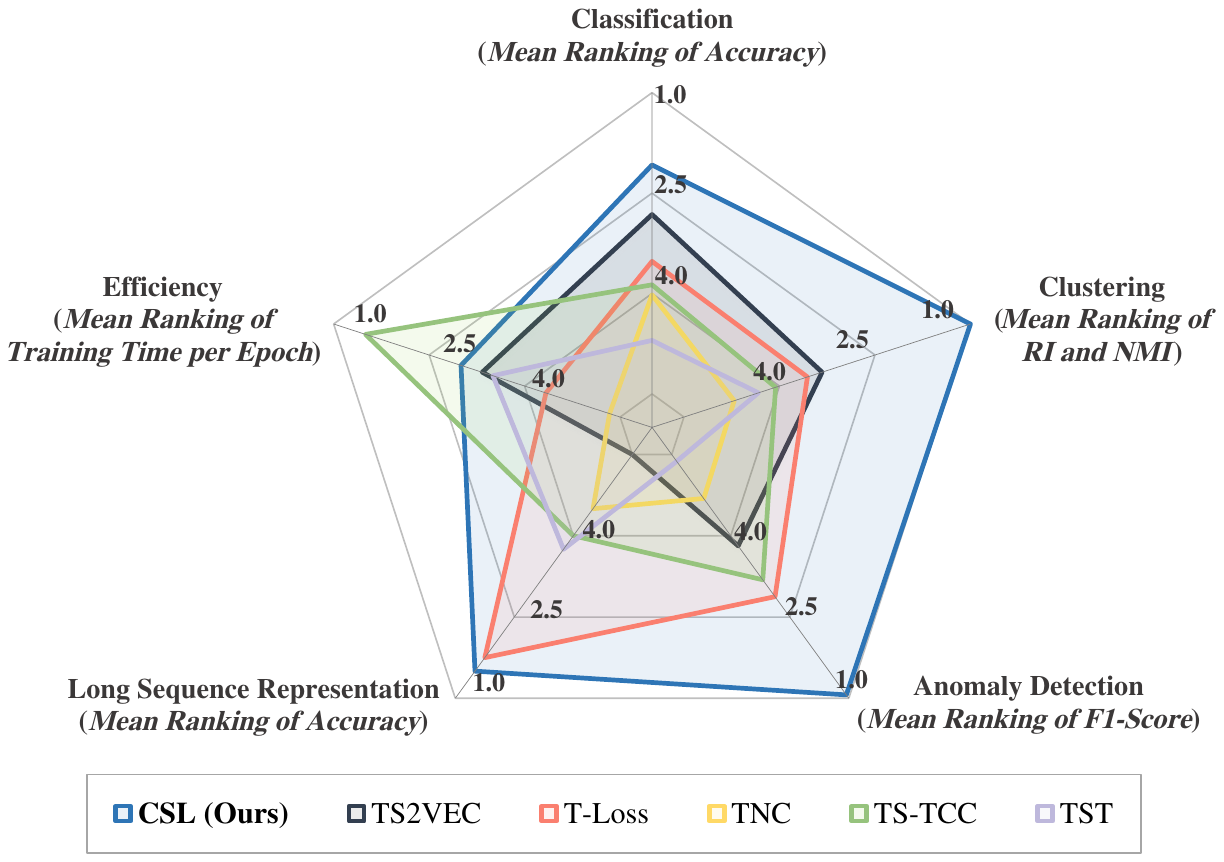}
    \caption{Overall performance of CSL against the competitors (smaller is better) regarding classificaton, clustering, anomaly detection, long sequence representation and training efficiency. See $\S$5.2, $\S$5.7 and $\S$5.8 in~\cite{liang2023shapelet} for the details.}
    \label{fig:performance}
\end{figure}

To address the above issues, we have proposed \myem{\underline{C}ontrastive \underline{S}hapelet \underline{L}earning (CSL)}~\cite{liang2023shapelet}, a brand new unsupervised representation learning framework for multivariate (and also univariate) time series. To the best of our knowledge, \myem{CSL is the first general-purpose URL method based on shapelet}, an interpretable pattern specifically designed for time series that represents the discriminative subsequence. Unlike traditional approaches that learn the shapelets for a specific analysis task, such as time series classification~\cite{icde22shapelets,liang2021efficient} or clustering~\cite{u-shapelets,USSL}, the shapelets of the proposed CSL are learned with unsupervised contrastive learning, which has shown superiority in many downstream analysis tasks~\cite{liang2023shapelet}, including classification, clustering, segment-level anomaly detection, and long time series representation. We summarize the performance of CSL against the competing methods in Figure~\ref{fig:performance}, which is extensively evaluated using 38 datasets from various real-world scenarios~\cite{liang2023shapelet}. 

In this paper, \myem{we demonstrate \texttt{TimeCSL}, a novel system that makes full use of CSL to achieve explorable time series analysis for various tasks in a unified process.} \texttt{TimeCSL} includes an end-to-end \myem{unified pipeline} that first learns the general shapelets of multiple scales and (dis)similarity metrics without any annotation by running the CSL algorithm~\cite{liang2023shapelet}. Then, it addresses different time series analysis tasks by building arbitrary task-oriented analyzers (e.g., SVM for classification and K-Means for clustering) on top of the general-purpose shapelet-based features. The pipeline has shown superior performance compared to that of the complex task-specific approaches, and significantly outperforms the traditional supervised methods when there are few available labels. We refer the interested readers to our research paper~\cite{liang2023shapelet} for more details.

\texttt{TimeCSL} provides flexible and intuitive \myem{visual exploration} of the raw time series, the learned shapelets, and the shapelet-based time series representation, offering a useful tool for interpreting the analysis results. Users can experiment with the system using their own data to explore the learned shapelet-based features, which are usually more insightful and intuitive-to-understand than the complex raw time series. This ``explorable'' analysis can help to explain the decisions made by the task-oriented analyzer.

\section{The \texttt{TimeCSL} System}

\begin{figure}
    \centering
    \includegraphics[width=\linewidth]{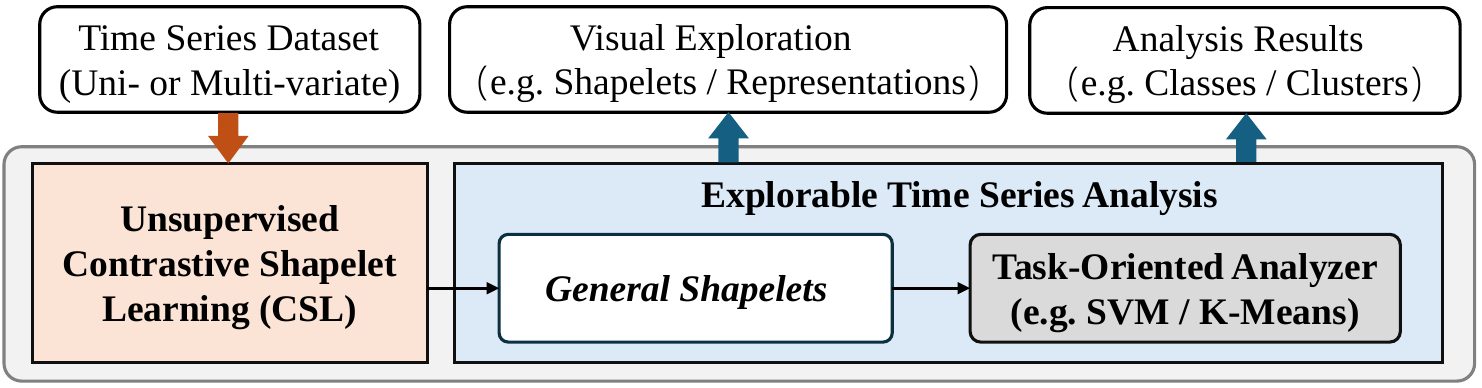}
    \caption{The \texttt{TimeCSL} pipeline.}
    \label{fig:pipeline}
\end{figure}

As depicted in Figure~\ref{fig:pipeline},  \texttt{TimeCSL} is comprised of two components, including \textit{Unsupervised Contrastive Shapelet Learning}  and \textit{Explorable Time Series Analysis}. These components work as follows.

\subsection{Unsupervised Contrastive Shapelet Learning}

The goal of this component is to learn general shapelets from the training time series, which transforms the raw time series into shapelet-based representation to facilitate different downstream analysis tasks. This is achieved using our proposed CSL method~\cite{liang2023shapelet}. 

Given a dataset containing $N$ time series as $\boldsymbol{X} = \{\boldsymbol{x}_1, \boldsymbol{x}_2, ..., \boldsymbol{x}_N\} \in \mathbb{R}^{N \times D \times T}$, where each time series $\xv_i \in \mathbb{R}^{D \times T}$ has $D$ variables ($D \geq 1$) and $T$ observations ordered by time, CSL embeds $\xv_i$ into the shapelet-based representation $\zv_i \in \mathbb{R}^{D_{repr}}$ using the proposed \textit{Shapelet Transformer} $f$, i.e. $\zv_i=f(\xv_i)$, where $f$ contains the learnable shapelets of various (dis)similarity measures and lengths (a.k.a. scales). CSL learns $f$ using an unsupervised contrastive learning algorithm, which iteratively optimizes the proposed \textit{Multi-Grained Contrasting} and \textit{Multi-Scale Alignment} objectives in an end-to-end manner with stochastic gradient descent.     

Using the Shapelet Transformer $f$ (i.e. all the shapelets) learned by CSL, the \texttt{TimeCSL} system transforms all input time series into the shapelet-based features as $\zv_i = f(\xv_i)$, and performs the downstream analysis tasks on top of the representation $\zv_i$. It is noteworthy that $\zv_i$ represents the (dis)similarity (e.g., the minimum Euclidean norm or the maximum cosine similarity) between the subsequences of $\xv_i$ and each of the shapelets, and therefore is fully interpretable and explainable. 

\subsection{Explorable Time Series Analysis}

By making full use of the general-purpose and explainable shapelet-based features learned by CSL, this component not only offers a unified and flexible way to perform different time series analysis tasks (e.g. classification, clustering, and anomaly detection),  but also the intuitive visual exploration of the raw time series, the learned shapelets, and the shapelet-based representation, so that the users can gain useful insights into their data to understand the decision basis of the analysis results (e.g. the predicted classes or clusters).

\textbf{Task solving.} As mentioned above, \texttt{TimeCSL} solves all different time series analysis tasks using the shapelet-based representation learned by CSL. This is achieved by building a task-oriented analyzer (e.g., SVM for classification or K-Means for clustering) that takes the shapelet-based feature vector $\zv_i$ as input and outputs the corresponding analysis results (e.g., classes or clusters). \texttt{TimeCSL} provides two modes to build the analyzer, including the \textit{freezing mode} and the \textit{fine-tuning mode}, which differ in whether to fine-tune the parameters of the Shapelet Transformer $f$ that is pre-trained by the Unsupervised Contrastive Shapelet Learning component. 

\begin{figure*}
    \centering
\subcaptionbox{\centering A raw time series.
\label{subfig:series}}
{\includegraphics[width=.44\linewidth]{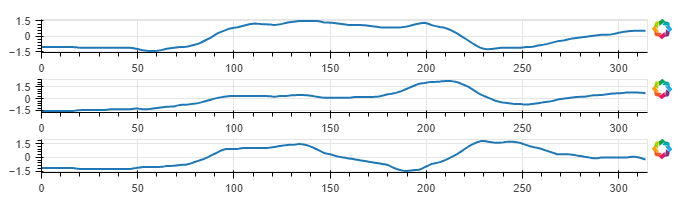}}
\subcaptionbox{\centering Matching between selected shapelet and time series.
    \label{subfig:matching}}
{\includegraphics[width=.46\linewidth]{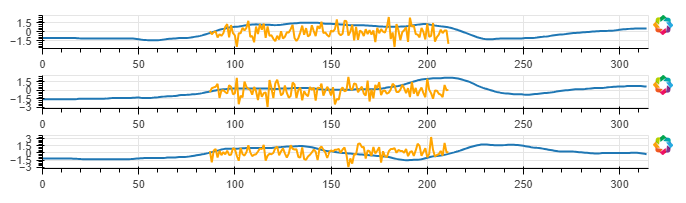}}

\vspace{2ex}
\subcaptionbox{\centering Learned shapelets grouped by length and (dis)similarity metric.
    \label{subfig:shapelet}}
{\includegraphics[width=.3\linewidth]{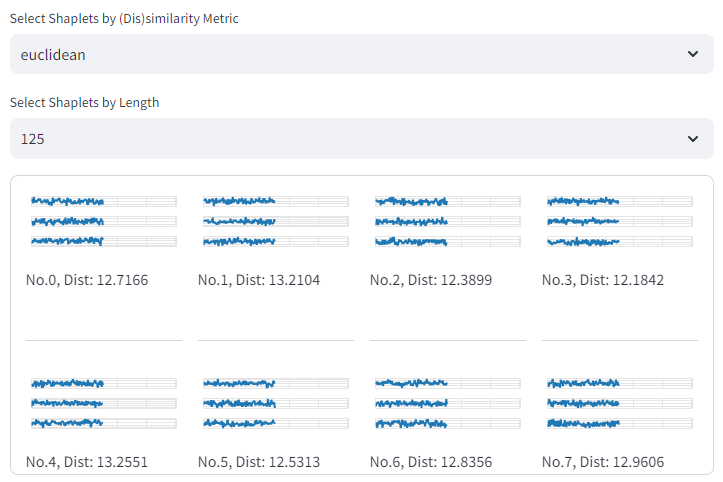}}
\subcaptionbox{\centering Shapelet-based representation of a dataset shown in a tabular form.
    \label{subfig:tabular}}
{\includegraphics[width=.33\linewidth]{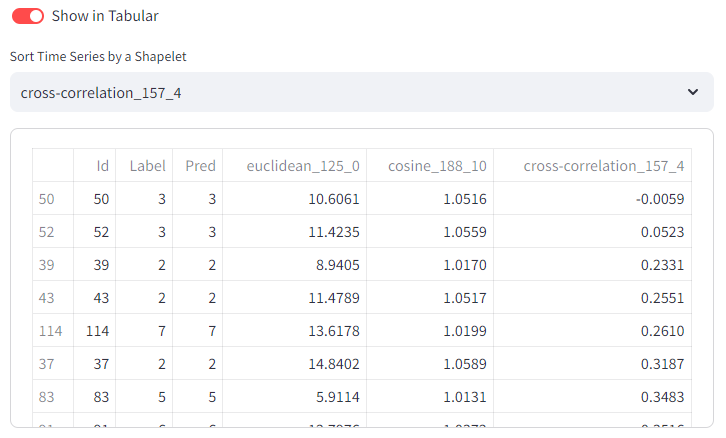}}
\subcaptionbox{\centering Shapelet-based representation of a dataset visualized using two-dimensional t-SNE~\cite{tSNE}.
    \label{subfig:tsne}}
{\includegraphics[width=.35\linewidth]{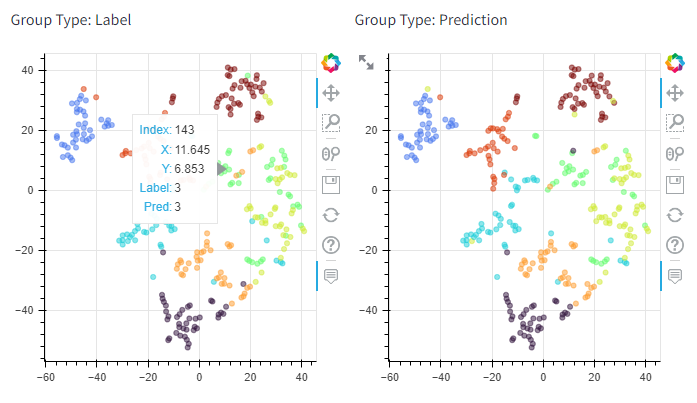}}
    \caption{The \texttt{TimeCSL} interface.}
    \label{fig:GUIs}
   \vspace{2ex}
\end{figure*}

$\bullet$ \underline{\textit{Freezing mode.}} In this basic mode, the task-oriented analyzer is built by directly using the pre-trained Shapelet Transformer $f$ to extract the general-purpose shapelet-based features, while the parameters of $f$ are kept unchanged during the building. Therefore, any standard analyzer (e.g. the many popular classifiers such as SVM, logistic regression, GBDT, etc) can be seamlessly integrated to facilitate different application scenarios of the users.   

$\bullet$ \underline{\textit{Fine-tuning mode.}} This is an advanced mode that allows users to fine-tune the parameters of the pre-trained Shapelet Transformer $f$ for specific tasks to further improve the accuracy of the analysis. In this mode, \texttt{TimeCSL} uses a task-specific neural network $g$ (e.g., a linear classifier) as the analyzer, which is appended to the pre-trained $f$ to map the shapelet-based features to the analysis results, i.e., $\hat{y}_i = g(\zv_i) = g(f(x_i))$ where $\hat{y}_i$ is a prediction (e.g. a predicted class) for the time series $\xv_i$. It simultaneously adjusts the parameters of $f$ and $g$ by minimizing the task-oriented loss function (e.g. the cross-entropy loss widely used for classification) using the backpropagation algorithm~\cite{rumelhart1986learning-bp}. Afterwards, the fine-tuned $f$ and $g$ are used for extracting the shapelet-based features and predicting the analysis results, respectively.

The fine-tuning mode of \texttt{TimeCSL} is \myem{especially effective in the semi-supervised scenarios}, where only a small portion of the data is annotated. Users can pre-train the Shapelet Transformer $f$ through Unsupervised Contrastive Shapelet Learning to make full use of all time series. Subsequently, they only need to fine-tune $f$ and $g$ using the available labeled data to achieve competitive performance. For example, our study on a real-world gesture recognition problem (see $\S$5.5 in~\cite{liang2023shapelet}) shows that, when the proportion of labeled data is less than 20\%, the fine-tuned method (using the Shapelet Transformer $f$ and a linear layer $g$) that takes advantage of the unsupervised pre-training of CSL, is 7\% to 10\% more accurate than the state-of-the-art customized time series classification approach.

\textbf{Visual exploration.} \texttt{TimeCSL} provides the visual exploration of the following two types of data for gaining insight into the time series and understanding the analysis results.

$\bullet$ \underline{\textit{Time series and shapelets.}} Since the shapelets represent the salient approximate subsequences of the raw time series which are of the same number of variates, \texttt{TimeCSL} shows both the raw time series and the learned shapelets in the line chart, where each line indicates the values of one variate.

$\bullet$ \underline{\textit{Shapelet-based features.}} By definition, a shapelet-based feature represents the (dis)similarity (e.g., Euclidean norm) between a shapelet and the most similar subsequence of a time series. Therefore, given a time series and a shapelet, \texttt{TimeCSL} visualizes them, matches the shapelet with its most similar subsequence, and displays the feature value measured by the corresponding (dis)similarity metric. In addition to this instance-level exploration, \texttt{TimeCSL} can also show the shapelet-based representation of a dataset in a tabular form, allowing flexible sorting of the time series according to each of the shapelets (equivalent to the features), and visualize the high-dimensional time series representation in a two-dimensional graph using t-SNE~\cite{tSNE}.   

The user can interact with \texttt{TimeCSL} to iteratively use the features extracted with different subsets of the shapelets in the explorable time series analysis. Given the interpretability of the shapelet-based representation, it is feasible to gain an understanding of the factors that influence the analysis results.

\section{Demonstration}
The demonstration\footnote{\url{https://youtu.be/hjdKy4Zr2Zw}} illustrates two key capabilities of the \texttt{TimeCSL} system. Firstly, it showcases how the user performs different time series analysis tasks using the unified pipeline. Secondly, it demonstrates how users gain insight into their time series by exploring the interpretable shapelet-based representation, which helps to understand the underlying decision basis of the analysis results. 

We prepare the UEA archive~\cite{UEA} for the audience to interact with our system, which contains 30 time series datasets from various applications. Users can also analyze and explore their own data. 

Generally, a user takes the following four steps to achieve explorable time series analysis using the \texttt{TimeCSL} system.

\textbf{Step 1: Configuring the Shapelet Transformer.} A user can flexibly configure the Shapelet Transformer to decide the number, length, and (dis)similarity metric of the shapelets to be learned, or directly start with our recommended configuration that uses Euclidean norm, cosine similarity, and cross-correlation function to measure the (dis)similarity, and adaptively sets the number and lengths of the shapelets based on the time series (see $\S$4.2 in~\cite{liang2023shapelet}).

\textbf{Step 2: Unsupervised Contrastive Learning of the General Shapelets.} After configuration, a user clicks on the ``Run CSL'' button to begin the unsupervised contrastive learning. Then, \texttt{TimeCSL} automatically learns the general and interpretable shapelets by running the CSL algorithm. A user can diagnose the model performance through the learning curve plotted in our GUI that records the training or validation loss over time.    

\textbf{Step 3: Performing the Analysis Tasks.}
\texttt{TimeCSL} transforms the time series of different lengths and numbers of variates into a unified vector representation using the learned shapelets. Therefore, a user can easily perform an analysis task by specifying a mode and a task-oriented analyzer. \texttt{TimeCSL} has integrated several commonly used analyzers. In the freezing mode, to name only a few, it integrates SVM, K-Means, and Isolation Forest using scikit-learn\footnote{\url{https://scikit-learn.org/stable/}}, for classification, clustering, and anomaly detection, respectively. It also implements a linear classifier using PyTorch\footnote{\url{https://pytorch.org/}} to deal with the (semi-supervised) classification problem in the fine-tuning mode.  Using the GUI, a user can select one of the tasks, modes, and analyzers, and then click ``Run Analyzer'' to get the analysis results.

\textbf{Step 4: Exploring the Data.} Figure~\ref{fig:GUIs} shows part of the GUI for analyzing the time series of the prepared UWaveGestureLibrary dataset with the default configuration of the system. A user can explore the time series (Figure~\ref{subfig:series}) and the learned shapelets (Figure~\ref{subfig:shapelet}). To see the feature of a time series extracted using a specific shapelet, a user only needs to select them and click on the ``Match'' button. \texttt{TimeCSL} will visually match the shapelet with the most similar subsequence of the time series to illustrate this shapelet-based feature (Figure~\ref{subfig:matching}). \texttt{TimeCSL} also provides two types of exploration of the high-dimensional shapelet-based representation of the dataset. A user can select a set of interested shapelets, and click ``Show in Tabular'' to see the features extracted using these shapelets in a table and sort the time series according to each of the shapelets or features (Figure~\ref{subfig:tabular}). The user can also click ``Show in t-SNE'' to investigate these features in a two-dimensional t-SNE representation, where the points are clickable so that the user can see details about the corresponding time series (Figure~\ref{subfig:tsne}). 

A user can redo the analysis task (i.e., Step 3) using the selected shapelets of interest to explore their influence on the analysis results. For example, if a user runs an SVM analyzer on UWaveGestureLibrary by selecting the shapelets of length 31 learned as above, s/he can get a classification accuracy of 0.75. If the user restarts Step 3 using the learned shapelets of larger lengths,  the resulting accuracy scores are, to name a few, 0.85 for length 97 and 0.89 for length 188, which is even comparable to the accuracy of 0.91 that is achieved by using all the learned shapelets. The user can infer from the results that, in this scenario, the categories of the time series can be better distinguished by focusing on longer salient subsequences.

\begin{acks}
We thank the anonymous reviewers for their time and the valuable
comments. This paper was supported by the NSFC grant (62232005, 62202126), the National Key Research and Development Program of China (2021YFB3300502), and the Postdoctoral Fellowship Program of CPSF (GZC20233457).
\end{acks}
%\clearpage

%\balance
\bibliographystyle{ACM-Reference-Format}
\bibliography{ref-demo}

\end{document}